\documentclass{article}
\PassOptionsToPackage{numbers,sort&compress}{natbib}
\usepackage[final,sglblindworkshop]{neurips_2026}
\workshoptitle{BeNTo: Beyond Next-Token Prediction --- Diffusion \& Flow Models for Next-Generation Decoding}
\usepackage[utf8]{inputenc}
\usepackage[T1]{fontenc}
\usepackage{amsmath,amssymb}
\usepackage{graphicx,booktabs,array}
\usepackage{xcolor}
\usepackage{needspace}
\usepackage{listings,textcomp}
\usepackage{flafter}
\usepackage{microtype}
\usepackage[hidelinks]{hyperref}
\usepackage{url}
\hypersetup{pdftitle={Two Clocks in Diffusion MLLMs: When Answers Stabilize Before Rationales Unfold},pdfauthor={Keuntae Kim and Yong Suk Choi},pdfsubject={BeNTo 2026 workshop camera-ready paper}}
\newcommand{\ts}{\ensuremath{T_{\mathrm{stab}}}}
\newcommand{\tc}{\ensuremath{T_{\mathrm{commit}}}}
\newcommand{\tp}{\ensuremath{T_{\mathrm{PDM}}}}
\newcommand{\rp}{\ensuremath{r_{\mathrm{pos}}}}
\newcommand{\lav}{LaViDa}
\newcommand{\nemo}{Nemotron}
\newcommand{\dg}{DiffusionGemma}
\newcommand{\mthree}{M$^3$CoT}
\newcommand{\vs}{V*Bench}
\newcommand{\sq}{ScienceQA}

\title{Two Clocks in Diffusion MLLMs:\\When Answers Stabilize Before Rationales Unfold}
\author{%
  Keuntae Kim \qquad Yong Suk Choi\thanks{Corresponding author.}\\
  Department of Computer Science, Hanyang University, Seoul, Korea\\
  \texttt{\{ktkpv94,cys\}@hanyang.ac.kr}%
}
\begin{document}
\setcounter{footnote}{1}
\maketitle
\setcounter{footnote}{0}

\begin{abstract}
An answer candidate in a masked diffusion MLLM can stabilize while its rationale is still unfolding. We distinguish retrospective stabilization of the logged candidate from token commitment, and examine these two clocks relative to rationale generation. Analyzing our results across three visual question-answering benchmarks, we find that 89.4--98.1\% of the rationale-side canvas remains unwritten at stabilization in single-block, EOS-suppressed \lav{} runs. On \vs, reducing block length from 128 to 8 changes this fraction from 89.4\% to 1.7\%, together with answer coverage and the eligible observation window. Under EOS-enabled prompting, direct instructions improve \nemo's overall accuracy by 15.0 and 19.5 percentage points on \mthree{} and \sq, but reduce \lav/\vs{} accuracy by 11.0 points. A symmetric decomposition associates the larger absolute component of each change with coverage rather than conditional accuracy. Matched-canvas image ablations measure visual sensitivity alongside answer stabilization, separating the two temporal readouts. Together, these measurements distinguish answer stabilization, rationale unfolding, and visual sensitivity, and identify coverage as the larger component of the prompting differences.
\end{abstract}

\section{Introduction}
\label{sec:intro}
A rationale printed before an answer appears to document the steps that produced it. Chain-of-thought (CoT) prompting can improve task performance~\citep{wei2022}, but the order and content of an explanation need not faithfully reveal the computation behind an answer~\citep{turpin2023,lanham2023}. In masked diffusion generation, this distinction is especially concrete: a token displayed late in the final response may be predicted, or even permanently written, before tokens displayed earlier.

Iterative masked decoding exposes a useful observation point. At a still-masked answer position, the decoder has a candidate token even when its scheduler does not yet select that position for commitment~\citep{sahoo2024,li2025prophet}. We distinguish two events: \ts, when the \emph{logged candidate} first matches the final answer and never subsequently departs from it on observable steps, and \tc, when that answer position becomes permanent. Stabilization is computed retrospectively from the recorded trajectory.

\begin{figure}[t]
\centering
\includegraphics[width=\linewidth]{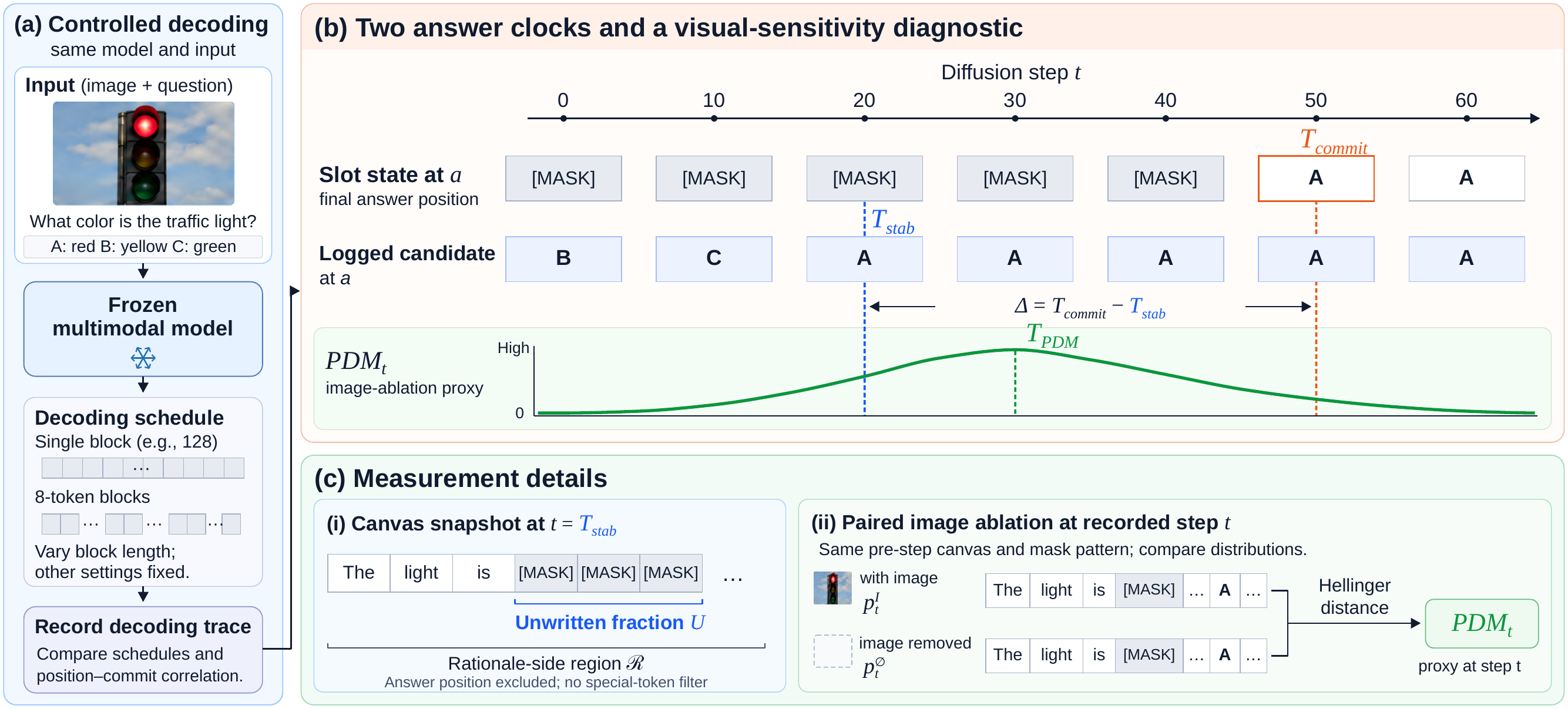}
\caption{\textbf{Overview of the measurement protocol.} (a) Hold the model and input fixed, vary the block schedule, and record the decoding trace. (b) Distinguish stabilization of the logged answer candidate, \ts, from permanent token commitment, \tc; the image-ablation peak \tp{} is a separate diagnostic. (c-i) Measure the unwritten fraction $U$ of the rationale-side region $\mathcal R$ at \ts. (c-ii) Compare image-present and image-removed token distributions on the same pre-step canvas and mask pattern. The PDM readout region is defined in Section~\ref{sec:pdm_method}. The trace and curve are schematic illustrations. After commitment, a logged candidate may be copied from the canvas. Here $\mathrm{PDM}_t=D_t$ in Equation~\ref{eq:pdm}.}
\label{fig:overview}
\end{figure}

Early answer convergence and answer-first commitment are already documented~\citep{li2025prophet,yeom2026,kim2026thinking}. We examine when answers stabilize relative to the unfolding rationale, alongside visual sensitivity and valid-answer coverage. Here, \emph{unfolding} refers to the commitment of rationale-side tokens to the canvas. Figure~\ref{fig:overview} separates the two answer clocks from the visual-sensitivity diagnostic.

A short-block schedule can make the answer position observable only after earlier text is written. A decreasing image-ablation signal can reflect information already in the canvas. An accuracy gain can arise chiefly because more responses contain extractable answers. None of these distinctions is visible in the completed rationale alone.

Our analysis connects three observations. First, the \lav{} block control links a fall in the unwritten fraction from 89.4\% to 1.7\% to changed answer-slot eligibility and lower answer coverage. Second, matched-canvas image ablations give configuration-dependent peak locations that cannot be substituted for answer stabilization. Third, coverage contributes the larger absolute term in all five prompting comparisons, including the \lav{} accuracy loss. Building on earlier convergence and commitment-order studies, these results characterize how rationale generation relates to answer stability, visual sensitivity, and answer availability under the evaluated decoding policies.

\section{Related work}
\label{sec:related}
\paragraph{Early convergence and commitment order.}
\citet{li2025prophet} study early answer convergence and introduce Prophet, an early-commit decoder. Their main convergence analysis tracks the first ground-truth-token match on finally correct outputs; their appendix also examines changes on incorrect outputs. Our target is a stable suffix matching the output's own final token on eligible logged steps. We use this distinction between prediction and commitment to examine how multimodal rationales unfold under explicit scheduling and stopping conditions.

\citet{yeom2026} study answer-first commitment, finite-canvas backfill, and positional restrictions. They also delay a reserved answer position and compare against a timing-matched reasoning-token delay. Our additional focus is the relationship to image-ablation sensitivity and to valid-answer coverage.

\paragraph{Visual grounding and rationale faithfulness.}
\citet{favero2024} introduce visual Prompt Dependency Measure (PDM), including its Hellinger-distance form PDM-H, for autoregressive VLMs. Thinking Diffusion~\citep{kim2026thinking} applies visual PDM to diffusion MLLMs, reports early answer generation and weak early visual dependence, and proposes Position and Step Penalty (PSP) and Visual Reasoning Guidance (VRG). We use matched-canvas distribution comparisons to measure the relationship between visual-sensitivity peaks and answer stabilization.

ST-Veto~\citep{kim2026st} uses second-order Taylor residuals of token confidence and image-attention mass to veto and replace unmasking candidates. Its online confidence-stability criterion differs from our retrospective final-token stability, and its attention-based grounding signal differs from our image-ablation distribution comparison. Turpin et~al.\ \citep{turpin2023} show that explanations can omit input biases that influence answers; Lanham et~al.\ \citep{lanham2023} intervene on CoT content to measure answer dependence.

\paragraph{Multimodal diffusion systems.}
\lav~\citep{li2025lavida}, the \nemo{} diffusion vision-language path~\citep{fu2026}, and \dg~\citep{diffusiongemma2026} provide different decoding implementations. We analyze answer timing in \lav{}, prompting and visual sensitivity in \lav{} and \nemo{}, and an additional prompting comparison in \dg{}.

\section{Measurements}
\label{sec:method}
\subsection{Observable candidates and the two clocks}
Let $c_t[i]$ denote generation position $i$ after step $t$, starting from a masked canvas of length $L$. For the monotone paths studied here, a position, once committed, is not revised. Let $K_i$ be its recorded commitment step, with $K_i=\infty$ for positions not committed in the recorded trajectory. We use zero-indexed steps.

We localize the token containing the final answer letter at position $a$ and record its final token ID $y$. This is token-level stability: a change in whitespace or punctuation tokenization need not change the selected option. Let $\hat y_{t,a}$ be the candidate stored in the trace. The eligible set $\mathcal E_a$ contains only logged steps at which the answer-position confidence is nonnegative; a negative sentinel marks an unobserved position outside the active block. The implemented clocks are
\begin{equation}
 \begin{aligned}
 \ts&=\min\{t\in\mathcal E_a:\hat y_{s,a}=y\ \text{for every }s\in\mathcal E_a,\ s\geq t\},\\
 \tc&=K_a,\qquad \Delta=\tc-\ts.
 \end{aligned}
 \label{eq:clocks}
\end{equation}
The minimum is undefined when the eligible suffix does not end in $y$. Such traces cannot contribute a stabilization time. Restricting the scan to eligible steps avoids assuming that predictions are observable throughout the global decode.

At an active masked slot, the greedy candidate is the decoder's top choice. At an already visible slot, the decoding interface can copy the stored candidate from the canvas. The clocks track these recorded candidates and the permanent writes to the canvas.

\subsection{Rationale unfolding under a commitment schedule}
\label{sec:span}
For the timing analysis, we use an operational region $\mathcal R$: positions before $a$ for \lav{}, and positions after $a$ for answer-first output formats. Its unwritten fraction is
\begin{equation}
 U=\frac{1}{|\mathcal R|}\sum_{i\in\mathcal R}\mathbf 1[K_i>\ts]
   =1-F_{\mathcal R}(\ts),
 \label{eq:u}
\end{equation}
Empty regions are excluded. There is no special-token filter. A token committed in the same step as stabilization counts as written. The reported $U$ is a mean of per-example fractions. The region $\mathcal R$ is the rationale-side canvas, including formatting tokens and all positions retained under EOS suppression. A large $U$ indicates that most of this region is still unwritten at answer stabilization.

Here $F_{\mathcal R}(t)=|\mathcal R|^{-1}\sum_{i\in\mathcal R}\mathbf 1[K_i\leq t]$ is the fraction of the region committed by step $t$. Thus $U$ is a derived measure of rationale-side completion, not an independent clock or independent evidence of early stabilization. Given $F_{\mathcal R}$, it adds no information beyond \ts; given only \ts{} and \tc{}, it locates stabilization relative to rationale unfolding. A fixed token-count schedule need not fix which rationale-side positions are committed. If their commitment times are fixed, $U$ is simply a nonincreasing transformation of \ts{} (Appendix~\ref{app:clock_details}).

We also average finite per-trace correlations $\rp=\mathrm{corr}(i,K_i)$ over committed positions. Near-one values indicate nearly positional commitment. Block length determines the eligible answer-slot steps $\mathcal E_a$ and the context available for prediction.

\subsection{Matched-canvas visual sensitivity}
\label{sec:pdm_method}
We use the Hellinger-distance form of PDM (PDM-H) introduced by \citet{favero2024}. Following its diffusion use in Thinking Diffusion~\citep{kim2026thinking}, we compare image-present and image-removed token distributions on the same pre-step text canvas and mask pattern. The readout region $\mathcal R_{\mathrm{vis}}$ uses a closing \texttt{</think>} boundary when present, otherwise the positional rule in Section~\ref{sec:span} (Appendix~\ref{app:pdm}). Then
\begin{equation}
 D_t=\frac{1}{|\mathcal R_{\mathrm{vis}}|}\sum_{i\in\mathcal R_{\mathrm{vis}}}
 \frac{\|\sqrt{p^I_{t,i}}-\sqrt{p^{\varnothing}_{t,i}}\|_2}{\sqrt{2}},
 \qquad \tp=\arg\max_t D_t.
 \label{eq:pdm}
\end{equation}
Each distance lies in $[0,1]$. The \lav{} path uses an additional forward pass with visual tokens replaced by padding in the unconditional prefix. The \nemo{} path replays recorded canvases with conditional and unconditional prefix caches.

Peak selection uses the first maximum of the finite readouts; a peak requires at least one such observation. Step $t$ labels the pre-step readout and subsequent commitment update. We report mean individual peak times, not the peak of an averaged curve. Image-derived text and visible-token copying can reduce sensitivity without removing visual information. The readout includes both masked and visible positions.

\subsection{Valid-answer coverage and accuracy}
\label{sec:coverage}
Let $C$ be the fraction of outputs yielding an option accepted by the scoring rule, $Q$ their conditional accuracy, and $A$ accuracy over all evaluated examples, counting extraction failures as incorrect. Then
\begin{equation}
 A=CQ.
 \label{eq:accuracy}
\end{equation}
We call $C$ \emph{valid-answer coverage}. An output is covered when the shared scoring rule extracts an accepted option. Missing answers and format or extraction failures count as uncovered. Prompting comparisons use this same scoring rule in both arms.

\section{Experimental setup}
\label{sec:setup}
We analyze \vs{} (191 examples), a fine-grained visual-search benchmark~\citep{wu2023}; \mthree{} (400 examples), a multimodal multi-step benchmark~\citep{chen2024}; and an image-containing \sq{} subset (400 examples)~\citep{lu2022}. Appendix~\ref{app:prompts} gives the instruction templates.

The primary \lav{} timing condition uses a 128-token canvas, 64 denoising steps, one block, greedy candidate selection, and EOS suppression. Its \vs{} control changes block length to 8. The \nemo{} timing comparisons disable the confidence early-exit shortcut and suppress EOS. Approximate dLLM-Cache is disabled in the traced settings, while fixed-prefix caches can still be used. Suppressing EOS permits full-canvas timing analysis.

Prompting results allow EOS and compare matched configured budgets within each model, using a shared scorer. For \nemo{} and \dg{}, the reasoning-requested prefix asks for \texttt{<think>}-tagged reasoning followed by an answer; the direct prefix requests only an option letter. Thus explanation instructions and response-format constraints change together. The \lav{} and \nemo{} comparisons use 128 tokens/64 steps; \dg{} retains its own sampler.

Timing localization maps the answer letter to its containing token. Output scoring separately strips echoed prompts, prioritizes explicit declarations, and uses a case-sensitive bare-letter fallback.

We analyze aggregate experimental results using the measurement definitions in Section~\ref{sec:method}. Timing means are conditional on valid localization; coverage and overall accuracy use the full evaluated set. We conducted the experiments on eight NVIDIA H100 GPUs. Appendices~\ref{app:measurements}--\ref{app:implementation} specify the extraction and aggregation rules, prompt templates, dataset interfaces, and decoding controls. We compare the results across these decoding configurations; Section~\ref{sec:correctness_stats} presents our correctness-stratified statistical tests.

\section{Results}
\label{sec:results}
\subsection{Answers stabilize while most rationale-side tokens remain unwritten}
\begin{figure}[htbp]
\centering
\includegraphics[width=\linewidth]{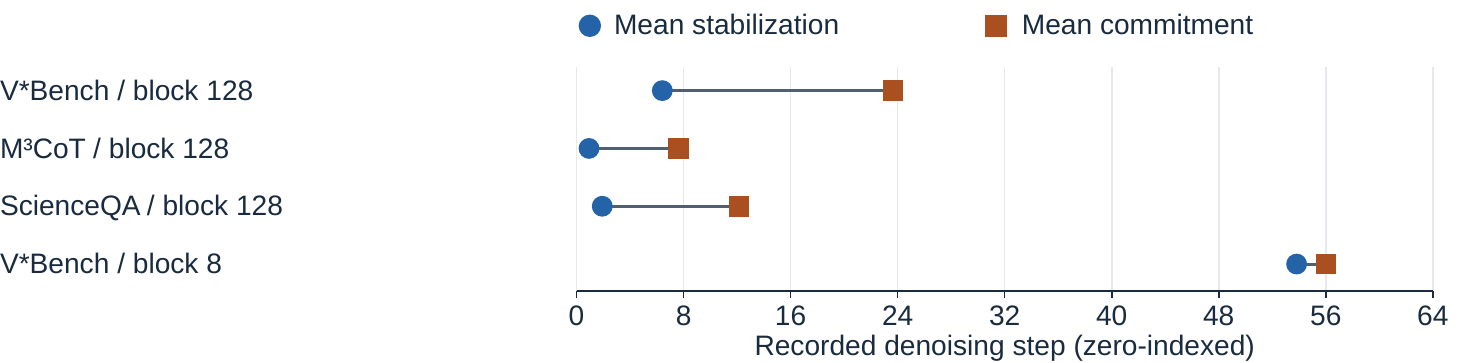}
\caption{\textbf{Recorded answer clocks and the block control.} Markers are mean recorded times, and segments are differences of means. All rows are \lav{} with EOS suppressed; the short-block condition also changes answer-slot eligibility.}
\label{fig:clocks}
\end{figure}

\begin{table}[htbp]
\caption{\textbf{\lav{} timing under single-block, EOS-suppressed decoding.} $N$ is the evaluation size; timing means are conditional on successful trace localization. $U$ measures the rationale-side canvas, not a semantic rationale annotation. Coverage and accuracy use all evaluated outputs.}
\label{tab:timing}
\centering\small
\begin{tabular}{lrrrrrrr}
\toprule
Data & $N$ & $\rp$ & $\ts$ & $\tc$ & $U$ (\%) & $C$ (\%) & $A$ (\%)\\
\midrule
\vs & 191 & 0.15 & 6.41 & 23.65 & 89.4 & 100.0 & 41.9\\
\mthree & 400 & 0.09 & 0.93 & 7.60 & 98.1 & 99.8 & 71.5\\
\sq & 400 & 0.19 & 1.92 & 12.14 & 96.2 & 97.5 & 62.5\\
\bottomrule
\end{tabular}
\end{table}
The displayed clock means differ by 17.24 steps on \vs, 6.67 on \mthree, and 10.22 on \sq{} (Table~\ref{tab:timing}). The corresponding unwritten fractions at stabilization are 89.4\%, 98.1\%, and 96.2\%. Figure~\ref{fig:clocks} visualizes the means and block control.

Timing means use traces with a localized answer and a nonempty rationale-side region. Coverage and overall accuracy use the full evaluation set.

The positional correlations, 0.09--0.19, show that commitment order differs from left-to-right generation. Stabilization tracks the final generated option for both correct and incorrect outputs. Section~\ref{sec:correctness_stats} reports correctness-stratified means and two-sided Mann--Whitney tests on \mthree{} and \sq{}, with high $U$ in both correctness groups.

\nemo{} has EOS-suppressed positional correlations of 0.93, 0.66, and 0.73 on \vs{}, \mthree{}, and \sq{}, respectively. These values show a stronger association between position and commitment time than the corresponding \lav{} values. The following sections report \nemo{} prompting and PDM results.

\subsection{Correctness-stratified statistical analysis}
\label{sec:correctness_stats}
We compare the per-example unwritten fraction $U$ at answer stabilization between correct and incorrect outputs on the two reasoning benchmarks, \mthree{} and \sq{}. Table~\ref{tab:correctness_stats} gives the group means and $p$-values. We partition examples by answer correctness and apply \texttt{scipy.stats.mannwhitneyu} with \texttt{alternative="two-sided"} to the two $U$ arrays. The displayed $p$-values are nominal, per-benchmark values.

\begin{table}[htbp]
\caption{\textbf{Correctness-stratified \lav{} results.} Mean $U$ is the rationale-side fraction still unwritten at stabilization, expressed as a percentage. The last column gives the two-sided Mann--Whitney $p$-value for the correct/incorrect comparison.}
\label{tab:correctness_stats}
\centering\small
\begin{tabular}{lrrr}
\toprule
Data & Correct: mean $U$ (\%) & Incorrect: mean $U$ (\%) & $p$ \\
\midrule
\mthree{} & 98.5 & 97.1 & 0.68 \\
\sq{} & 96.8 & 95.1 & 0.038 \\
\bottomrule
\end{tabular}
\end{table}

Both correctness groups have mean $U$ above 95\% on each benchmark. Our \sq{} analysis includes 398 analyzable examples and a 1.7-percentage-point difference between the group means. These comparisons place the high unwritten fractions in the context of answer correctness.

\subsection{The block control changes measurement and output coverage}
\begin{table}[htbp]
\caption{\textbf{Block-length control on \lav/\vs.} Both rows evaluate 191 examples with the same configured canvas and step budget. $Q$ is accuracy conditional on a valid extracted answer.}
\label{tab:block}
\centering\small
\begin{tabular}{rrrrrrrr}
\toprule
Block & $\rp$ & $\ts$ & $\tc$ & $U$ (\%) & $C$ (\%) & $A$ (\%) & $Q$ (\%)\\
\midrule
128 & 0.15 & 6.41 & 23.65 & 89.4 & 100.0 & 41.9 & 41.9\\
8 & 1.00 & 53.82 & 56.01 & 1.7 & 91.6 & 39.8 & 43.4\\
\bottomrule
\end{tabular}
\end{table}
Short blocks shift stabilization from step 6.41 to 53.82 and commitment from 23.65 to 56.01. The gap between the displayed means shrinks from 17.24 to 2.19 steps, while $U$ falls by 87.7 percentage points. The block policy delays access to the answer position and changes the context on which it is predicted.

Coverage drops by 8.4 points, overall accuracy by 2.1 points, and conditional accuracy rises by 1.5 points (Table~\ref{tab:block}). A more textually ordered decode is not uniformly better. This comparison reports configuration-level timing means alongside full-set coverage, capturing both the observed ordering and answer availability under each block policy.

The first eligible answer-slot step, $T_{\mathrm{first}}=\min\mathcal E_a$, separates two sources of delay: waiting for the slot to become observable and changes after it becomes observable. A short-block schedule can increase the first component even when a candidate stabilizes soon after observation begins. The reported global stabilization time includes both components and describes the complete decoding schedule.

\subsection{Visual-sensitivity summaries}
\begin{figure}[htbp]
\centering
\includegraphics[width=\linewidth]{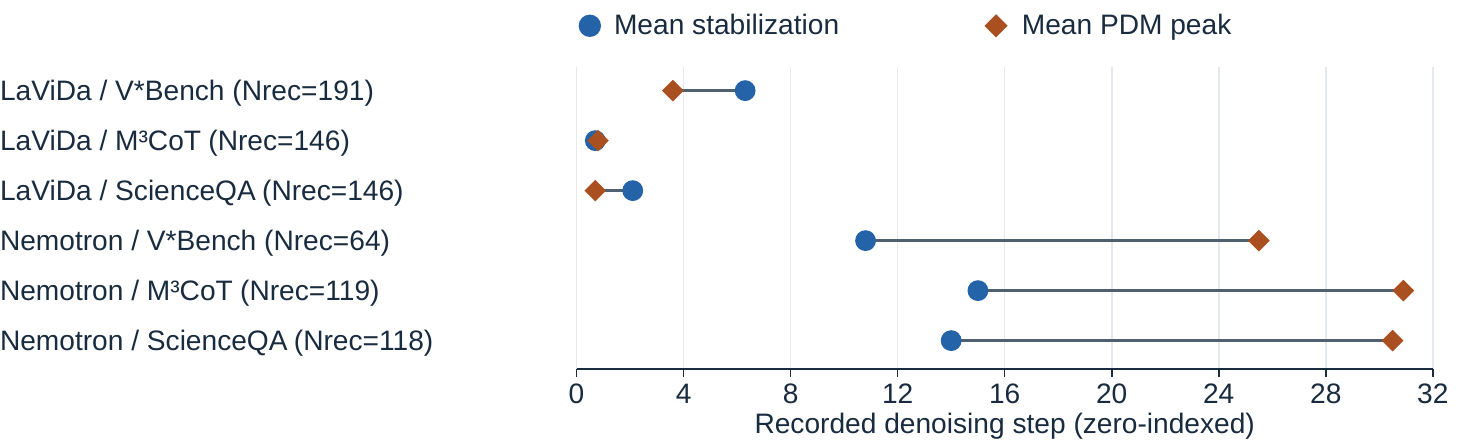}
\caption{\textbf{Visual-sensitivity timing.} Markers are mean individual stabilization and PDM peak steps. $N_{\rm rec}$ denotes the recorded subset size. Each row gives the two means for one model--dataset configuration.}
\label{fig:pdm}
\end{figure}
Our PDM analysis places \nemo's mean sensitivity peaks at steps 25.5, 30.9, and 30.5, versus 3.6, 0.8, and 0.7 for \lav{} (Figure~\ref{fig:pdm}). The associated stabilization means are 10.8, 15.0, and 14.0 for \nemo{}, and 6.3, 0.7, and 2.1 for \lav{}. These values come from the PDM evaluations listed in Appendix~\ref{app:pdm_values}.

Final sensitivity relative to mean individual peak sensitivity is 25\%, 13\%, and 12\% for \lav{}, versus 28\%, 47\%, and 45\% for \nemo. These quantities summarize mean peak locations and the image's relative influence at the final readout. They distinguish visual sensitivity from answer stability: each statistic measures a different property of the decoding trajectory. The peak locations describe the evaluated configurations, while the matched-canvas comparison measures the image's additional influence given the text already present.

\subsection{Direct prompting has model-dependent effects on answer coverage}
\begin{table}[htbp]
\caption{\textbf{EOS-enabled prompting comparisons.} Entries are reasoning-requested $\rightarrow$ direct (percent), with a shared scorer within each pair. $C$: valid-answer coverage; $A$: overall accuracy; $Q$: conditional accuracy. $N=191$ for \vs{} and $400$ for the other rows. \dg{} retains its own sampler.}
\label{tab:prompt}
\centering\small
\begin{tabular}{llrrr}
\toprule
Model & Data & $C$ & $A$ & $Q$\\
\midrule
\nemo & \vs & $90.6\to99.5$ & $47.6\to52.4$ & $52.6\to52.6$\\
\nemo & \mthree & $73.5\to88.8$ & $54.5\to69.5$ & $74.1\to78.3$\\
\nemo & \sq & $69.2\to90.2$ & $57.0\to76.5$ & $82.3\to84.8$\\
\dg & \vs & $84.3\to91.1$ & $53.9\to58.6$ & $64.0\to64.4$\\
\lav & \vs & $84.8\to66.5$ & $39.8\to28.8$ & $46.9\to43.3$\\
\bottomrule
\end{tabular}
\end{table}
For \nemo/\vs, $Q$ is 52.6\% in both arms at the reported precision, while $C$ rises from 90.6\% to 99.5\% and $A$ rises by 4.8 points. \dg/\vs{} shows a 6.8-point coverage gain, a 4.7-point overall-accuracy gain, and a 0.4-point conditional-accuracy change.

On \mthree{} and \sq, direct prompting increases \nemo's $A$ by 15.0 and 19.5 points, $C$ by 15.3 and 21.0, and $Q$ by 4.2 and 2.5. Both datasets therefore show increases in answer coverage and conditional accuracy, with the larger percentage-point change in coverage.

For \lav/\vs{}, direct prompting reduces all three rates: $C$ falls by 18.3 points, $Q$ by 3.6, and $A$ by 11.0. The direction of the prompting effect therefore differs across the evaluated models.

\subsection{An arithmetic decomposition of the prompting differences}
The product in Equation~\ref{eq:accuracy} permits a symmetric accounting of the change between reasoning-requested ($R$) and direct ($D$):
\begin{equation}
 A_D-A_R=
 \underbrace{(C_D-C_R)\frac{Q_D+Q_R}{2}}_{\text{coverage-associated term}}+
 \underbrace{(Q_D-Q_R)\frac{C_D+C_R}{2}}_{\text{conditional-accuracy-associated term}}.
 \label{eq:decomp}
\end{equation}
This identity follows by expansion and averages the two choices of reference arm. The coverage term combines the change in $C$ with the mean $Q$; the conditional-accuracy term combines the change in $Q$ with the mean $C$.

\begin{table}[htbp]
\caption{\textbf{Decomposition from the reported percentages.} Values are percentage points. The two terms are computed from rounded $C,Q$ and can sum slightly differently from the displayed $\Delta A$; the largest residual is 0.12 points.}
\label{tab:decomp}
\centering\small
\begin{tabular}{llrrr}
\toprule
Model & Data & $\Delta A$ & Coverage term & Conditional term\\
\midrule
\nemo & \vs & $+4.8$ & $+4.68$ & $0.00$\\
\nemo & \mthree & $+15.0$ & $+11.66$ & $+3.41$\\
\nemo & \sq & $+19.5$ & $+17.55$ & $+1.99$\\
\dg & \vs & $+4.7$ & $+4.37$ & $+0.35$\\
\lav & \vs & $-11.0$ & $-8.25$ & $-2.72$\\
\bottomrule
\end{tabular}
\end{table}
The coverage-associated term is larger in absolute magnitude in every row, including the \lav{} regression. For \nemo/\vs{}, the displayed $Q$ is unchanged and the conditional term is zero. In the other four rows, both terms have the same sign as the overall accuracy change.

The symmetric form averages two exact ways to expand the product difference. Using $Q_R$ as the reference gives $\Delta C\,Q_R+C_D\,\Delta Q$, while using $Q_D$ gives $\Delta C\,Q_D+C_R\,\Delta Q$, where $\Delta C=C_D-C_R$ and $\Delta Q=Q_D-Q_R$. Averaging these expressions assigns the interaction between the two rate changes equally to the two terms. Reversing the comparison reverses both signs while preserving their absolute magnitudes. The same accounting therefore applies to improvements and regressions.

For \nemo/\mthree{}, the 15.3-point coverage increase is weighted by the mean conditional accuracy of 76.2\%, giving 11.66 points. The 4.2-point conditional-accuracy increase is weighted by mean coverage of 81.15\%, giving 3.41 points. On \sq{}, the corresponding terms are 17.55 and 1.99 points. Thus the larger overall gain on \sq{} accompanies a larger coverage component, even though its conditional-accuracy increase is smaller than on \mthree{}. For \lav/\vs{}, the negative coverage term of $-8.25$ points and conditional term of $-2.72$ points jointly account for the accuracy loss.

\section{Discussion}
\label{sec:discussion}
\paragraph{Answer stability and rationale unfolding.}
The two clocks separate candidate stabilization from permanent token commitment. In single-block \lav{} decoding, the mean stabilization time precedes the mean commitment time on all three benchmarks, while 89.4--98.1\% of the rationale-side canvas remains unwritten at stabilization. The answer candidate is stable during a substantial portion of subsequent rationale generation.

The relationship between the clock gap and rationale progress is particularly clear when comparing \mthree{} and \sq{}. The difference of clock means is smaller on \mthree{} (6.67 versus 10.22 steps), yet its unwritten fraction is larger (98.1\% versus 96.2\%). The clock gap measures separation between two events at the answer position, whereas $U$ summarizes commitment across the rationale-side positions at stabilization. These readouts capture different aspects of the same trajectory. Equation~\ref{eq:u} connects them through the rationale-side commitment distribution: evaluating its cumulative fraction at \ts{} gives the written fraction, and taking the complement gives $U$. This relationship explains the role of $U$ alongside the two clocks in the timing tables.

\paragraph{Scheduling and visual sensitivity.}
The short-block control shifts both answer clocks later and reduces the unwritten fraction to 1.7\%. The PDM comparisons measure a separate temporal property: the step at which image removal changes token distributions most strongly. On \nemo{}, mean PDM peak times follow mean answer stabilization times on all three benchmarks. On \lav{}, the mean PDM peak precedes mean stabilization on \vs{} and \sq{}, while the two means are close on \mthree{}. PDM compares full vocabulary distributions, whereas stabilization follows one token identity. Changes in distributional sensitivity and persistence of the answer candidate are therefore compatible.

\paragraph{Coverage and prompting.}
Direct prompting raises \nemo{} and \dg{} accuracy and lowers \lav/\vs{} accuracy. In each comparison, the coverage-associated term is the larger absolute component of the accuracy change. Reporting $C$, $Q$, and $A$ together makes this pattern explicit: the prompting differences involve both accepted-answer frequency and correctness among accepted answers.

\section{Limitations}
\label{sec:limitations}
The block control exposes a trade-off between generation order and task performance. Reducing the block length lowers $U$ from 89.4\% to 1.7\%, while coverage falls from 100.0\% to 91.6\% and overall accuracy from 41.9\% to 39.8\%. More positional commitment therefore accompanies lower answer coverage and accuracy in this comparison. Direct prompting also has opposing effects across models: the accuracy gains on \nemo{} and \dg{} contrast with the 11.0-point loss on \lav/\vs{}.

\section{Conclusion}
\label{sec:conclusion}
In single-block, EOS-suppressed \lav{} decoding, answers stabilize before most rationale-side tokens are committed. Short blocks move both answer clocks later and sharply reduce the unwritten fraction. Image-ablation peaks occupy distinct mean step locations from answer stabilization. Direct prompting improves \nemo{} and \dg{} accuracy but reduces \lav/\vs{} accuracy, with coverage providing the larger arithmetic component in all five comparisons. Reading a rationale together with its commitment trace reveals two orderings: the order in which the explanation is presented and the order in which the model makes its output tokens permanent. The two clocks distinguish answer stability from commitment and connect both to the unfolding of the rationale.

\clearpage
\begin{ack}
This work was supported by the Institute of Information and communications Technology Planning and evaluation (IITP) grant (No.RS-2025-25422680, No. RS-2020-II201373), and the National Research Foundation of Korea (NRF) grant (No. RS-2025-00520618) funded by the Korean Government (MSIT).

The authors declare no competing interests.
\end{ack}



\clearpage
\appendix

\section{Supplementary measurement details}
\label{app:measurements}

\subsection{Answer extraction and token localization}
\label{app:extraction}
The scorer removes echoed prompts where applicable and prioritizes explicit answer declarations, parenthesized options, and line-initial letters, followed by a case-sensitive bare-capital fallback. Within each pattern it takes the last match. Extraction failures count as incorrect in overall accuracy and uncovered in valid-answer coverage.

The localizer joins individually decoded token pieces and maps an answer-letter character to its containing token: the last matching answer for \lav{}, the first for answer-first formats. Boxed answers take precedence over the other answer patterns. Tokenization, whitespace, or punctuation can change the tracked token ID without changing the extracted option. Scoring and trace localization are distinct operations.

\subsection{Clock computation and rationale-side region}
\label{app:clock_details}
Scanning eligible steps $\mathcal E_a$ backward identifies the longest suffix matching the final answer token $y$; its first step is $\ts$. An empty eligible set or a mismatch at the final eligible step makes stabilization undefined. Commitment $K_a$ is the first permanent write.

The timing region $\mathcal R$ comprises positions before the answer for \lav{}, and after it for answer-first formats. It excludes the answer and includes special tokens; empty regions are omitted. Negative commitment sentinels correspond to $K_i=\infty$, while $K_i=\ts$ counts as written. Averaging is over per-example fractions, not pooled tokens. Positional correlation averages finite values from traces with more than four committed positions.

\paragraph{What $U$ adds to the clocks.}
Equation~\ref{eq:u} expresses $U$ as the fraction of rationale-side commitment times later than \ts. For an illustrative four-position region, \ts$=2$ gives $U=1/2$ when its commitment times are $(1,2,3,4)$ and $U=1$ when they are $(3,4,5,6)$. Both traces can have \tc$=7$. This constructed example is not experimental data: it shows that the two answer clocks alone do not determine writing progress. Once the rationale-side commitment times are also given, $U$ is fully determined. Even under a fixed number of commits per step, allocation inside versus outside $\mathcal R$ can differ.

For a common valid example set $S$, the paired mean gap is
\begin{equation}
 \overline{\Delta}_S=\frac{1}{|S|}\sum_{j\in S}\left(T_{\mathrm{commit},j}-T_{\mathrm{stab},j}\right).
 \label{eq:paired_gap}
\end{equation}
It equals a difference of clock means when both means use $S$. Tables~\ref{tab:timing} and~\ref{tab:block} compare the recorded means separately. Their evaluation size $N$ describes the full task evaluation; metric-specific validity follows the definitions above.

\subsection{Visual-sensitivity readout and aggregation}
\label{app:pdm}
Step $t$ labels distributions on the same pre-update canvas and the commitment update they precede. The peak is the first finite maximizing readout. For PDM, we select positions before the token containing the first \texttt{</think>}, when present; otherwise we use the model-specific before/after-answer rule, clipped to the generation/readout length. This set is $\mathcal R_{\mathrm{vis}}$ in Equation~\ref{eq:pdm}. It includes formatting tokens; an answer inside the tag-delimited prefix is included in that prefix.

We average the individual peak times. Relative final sensitivity is
\begin{equation}
 \rho=100\,\frac{\overline{D_{\mathrm{final}}}}{\overline{\max_t D_t}}.
 \label{eq:pdm_ratio}
\end{equation}
This is a ratio of means. Visible text may already contain image-derived information, so matched-canvas ablation measures the image's additional influence at that canvas.

\clearpage
\section{Prompting and supplementary results}
\label{app:additional}

\subsection{Instruction templates}
\label{app:prompts}
For \nemo{} and \dg{}, we prepend the following instructions to the benchmark question and choices, then apply the model-specific image/text chat templates.

\paragraph{Reasoning-requested.}
\begin{quote}\small
You should first think about the reasoning process in the mind and then provide the user with the answer. The reasoning process is enclosed within \texttt{<think> </think>} tags, i.e.\ \texttt{<think>} reasoning process here \texttt{</think>} answer here
\end{quote}

\paragraph{Direct.}
\begin{quote}\small
Answer the question directly. Do not explain your reasoning and do not show any steps. Reply with only the letter of the correct option.
\end{quote}

The prefixes change both the requested explanation behavior and response-format constraints. The comparison uses a shared scorer for both instruction policies.

\subsection{Visual-sensitivity values}
\label{app:pdm_values}
Table~\ref{tab:pdm_detail} gives the numerical values underlying Figure~\ref{fig:pdm}. These are summaries of model-specific decoding paths. Peak steps are measured on the recorded denoising-step axis.

\begin{table}[htbp]
\caption{\textbf{PDM timing and relative final sensitivity.} $N_{\rm rec}$ is the recorded subset size. The final/peak column is $\rho$ from Equation~\ref{eq:pdm_ratio}. Stabilization means belong to these PDM evaluations, not the primary timing conditions in Table~\ref{tab:timing}.}
\label{tab:pdm_detail}
\centering\small
\begin{tabular}{llrrrr}
\toprule
Model & Data & $N_{\rm rec}$ & Mean $\ts$ & Mean $\tp$ & Final/peak (\%)\\
\midrule
\lav & \vs & 191 & 6.3 & 3.6 & 25\\
\lav & \mthree & 146 & 0.7 & 0.8 & 13\\
\lav & \sq & 146 & 2.1 & 0.7 & 12\\
\nemo & \vs & 64 & 10.8 & 25.5 & 28\\
\nemo & \mthree & 119 & 15.0 & 30.9 & 47\\
\nemo & \sq & 118 & 14.0 & 30.5 & 45\\
\bottomrule
\end{tabular}
\end{table}

\subsection{Accuracy decomposition and rounding}
\label{app:rounding}
Expanding Equation~\ref{eq:decomp} gives
\begin{align}
 &(C_D-C_R)(Q_D+Q_R)/2+(Q_D-Q_R)(C_D+C_R)/2\notag\\
 &\hspace{35mm}=C_DQ_D-C_RQ_R.
\end{align}
For \nemo/\mthree{}, the two terms are 11.6586 and 3.4083 percentage points. Their sum is 15.0669, compared with the displayed change of 15.0. Across all five rows, the largest absolute residual from independently rounded rates is 0.119 points.

\clearpage
\section{Implementation and reporting details}
\label{app:implementation}

\subsection{Data, input construction, and decoding}
\label{app:data_access}
We use the \texttt{test} split from each of the following dataset repositories. Section~\ref{sec:setup} gives the evaluated sample counts.
\begin{center}\small
\begin{tabular}{@{}ll@{}}
\toprule
Benchmark & Repository identifier \\
\midrule
\vs{} & \texttt{lmms-lab/vstar-bench} \\
\mthree{} & \texttt{LightChen2333/M3CoT} \\
\sq{} & \texttt{derek-thomas/ScienceQA} \\
\bottomrule
\end{tabular}
\end{center}

\label{app:runner_details}
For \vs{}, we use the dataset's question text. For \mthree/\sq{}, we combine optional context or hint, the question, and labeled choices; \sq{} answer indices map to option letters. The \sq{} interface filters for image-bearing rows. The \nemo{} input path uses RGB JPEG images and \texttt{process\_messages}; \dg{} uses an RGB image and its processor's chat template.

\label{app:software}
The \lav{} wrapper exposes token/step budgets, block length, schedule, remasking, temperature, seed, EOS, and cache controls. The \nemo{} runner uses BF16, with an early-exit threshold separate from its EOS policy. The \dg{} runner exposes its own temperature schedule and stopping controls. Section~\ref{sec:setup} gives the settings for each comparison.

\subsection{Compute resources}
\label{app:compute}
The experiments used eight NVIDIA H100 GPUs. The principal \lav/\nemo{} comparisons use a configured 128-token canvas and 64 denoising steps, with block and EOS conditions stated in Section~\ref{sec:setup}.

\subsection{Broader impacts}
\label{app:impacts}
The analysis separates answer stability, permanent commitment, visual sensitivity, and valid-answer coverage. Reporting these quantities together makes the generation process easier to interpret. The observed losses in the short-block control and the \lav{} direct-prompting comparison also document the performance trade-offs of changing the decoding schedule or instruction policy.

\subsection{Asset attribution and terms}
\label{app:licenses}
The original model and benchmark creators are cited in Sections~\ref{sec:related}--\ref{sec:setup}. The \href{https://github.com/jacklishufan/LaViDa/blob/main/LICENSE}{LaViDa code} uses Apache-2.0; the \href{https://huggingface.co/nvidia/Nemotron-Labs-Diffusion-VLM-8B}{Nemotron VLM card} specifies NSCLv1; and the \href{https://huggingface.co/google/diffusiongemma-26B-A4B-it}{DiffusionGemma card} specifies Apache-2.0. The \href{https://github.com/penghao-wu/vstar/blob/main/LICENSE}{V* project code} uses MIT, separately from rights to its photographs. The download section of the \href{https://lightchen233.github.io/m3cot.github.io/}{M$^3$CoT project} identifies the dataset license as CC BY-NC-SA. The \href{https://github.com/lupantech/ScienceQA/blob/main/README.md}{ScienceQA project} specifies CC BY-NC-SA 4.0 for the dataset. Third-party model, image, and dataset rights remain with their respective owners.


\clearpage

\section*{NeurIPS Paper Checklist}

\begin{enumerate}

\Needspace{11\baselineskip}
\item {\bf Claims}
    \item[] Question: Do the main claims made in the abstract and introduction accurately reflect the paper's contributions and scope?
    \item[] Answer: \answerYes{}
    \item[] Justification: The abstract and Section~\ref{sec:intro} state the measurement-focused contribution and the scope of the aggregate comparisons.

\Needspace{11\baselineskip}
\item {\bf Limitations}
    \item[] Question: Does the paper discuss the limitations of the work performed by the authors?
    \item[] Answer: \answerYes{}
    \item[] Justification: Section~\ref{sec:limitations} discusses the coverage and accuracy losses in the block control and the model-dependent effects of direct prompting.

\Needspace{11\baselineskip}
\item {\bf Theory assumptions and proofs}
    \item[] Question: For each theoretical result, does the paper provide the full set of assumptions and a complete (and correct) proof?
    \item[] Answer: \answerNA{}
    \item[] Justification: This is an empirical analysis; Appendix~\ref{app:rounding} gives the algebra for the accuracy decomposition.

\Needspace{11\baselineskip}
\item {\bf Experimental result reproducibility}
    \item[] Question: Does the paper fully disclose all the information needed to reproduce the main experimental results of the paper to the extent that it affects the main claims and/or conclusions of the paper (regardless of whether the code and data are provided or not)?
    \item[] Answer: \answerYes{}
    \item[] Justification: For the aggregate comparisons, Section~\ref{sec:method} and Appendix~\ref{app:measurements} specify the calculations, and Tables~\ref{tab:timing}--\ref{tab:pdm_detail} provide their numerical inputs.

\Needspace{11\baselineskip}
\item {\bf Open access to data and code}
    \item[] Question: Does the paper provide open access to the data and code, with sufficient instructions to faithfully reproduce the main experimental results, as described in supplemental material?
    \item[] Answer: \answerNo{}
    \item[] Justification: This paper does not include a public repository for the model-execution and analysis code.

\Needspace{11\baselineskip}
\item {\bf Experimental setting/details}
    \item[] Question: Does the paper specify all the training and test details (e.g., data splits, hyperparameters, how they were chosen, type of optimizer) necessary to understand the results?
    \item[] Answer: \answerYes{}
    \item[] Justification: Section~\ref{sec:setup} and Appendices~\ref{app:prompts} and~\ref{app:implementation} describe the datasets, inputs, decoding conditions, and scoring rules.

\Needspace{11\baselineskip}
\item {\bf Experiment statistical significance}
    \item[] Question: Does the paper report error bars suitably and correctly defined or other appropriate information about the statistical significance of the experiments?
    \item[] Answer: \answerNo{}
    \item[] Justification: The main timing and prompting comparisons present point estimates. Section~\ref{sec:correctness_stats} reports the two-sided Mann--Whitney tests for the correctness-stratified comparison.

\Needspace{11\baselineskip}
\item {\bf Experiments compute resources}
    \item[] Question: For each experiment, does the paper provide sufficient information on the computer resources (type of compute workers, memory, time of execution) needed to reproduce the experiments?
    \item[] Answer: \answerYes{}
    \item[] Justification: Section~\ref{sec:setup} and Appendix~\ref{app:compute} specify the use of eight NVIDIA H100 GPUs and the configured token, step, block, and EOS settings.

\Needspace{11\baselineskip}
\item {\bf Code of ethics}
    \item[] Question: Does the research conducted in the paper conform, in every respect, with the NeurIPS Code of Ethics \url{https://neurips.cc/public/EthicsGuidelines}?
    \item[] Answer: \answerYes{}
    \item[] Justification: The authors affirm that the research conforms to the NeurIPS Code of Ethics.

\Needspace{11\baselineskip}
\item {\bf Broader impacts}
    \item[] Question: Does the paper discuss both potential positive societal impacts and negative societal impacts of the work performed?
    \item[] Answer: \answerYes{}
    \item[] Justification: Appendix~\ref{app:impacts} discusses interpretation of the generation process and the observed performance trade-offs.

\Needspace{11\baselineskip}
\item {\bf Safeguards}
    \item[] Question: Does the paper describe safeguards that have been put in place for responsible release of data or models that have a high risk for misuse (e.g., pre-trained language models, image generators, or scraped datasets)?
    \item[] Answer: \answerNA{}
    \item[] Justification: This submission does not release pretrained weights, newly scraped data, or other high-risk generative assets.

\Needspace{11\baselineskip}
\item {\bf Licenses for existing assets}
    \item[] Question: Are the creators or original owners of assets (e.g., code, data, models), used in the paper, properly credited and are the license and terms of use explicitly mentioned and properly respected?
    \item[] Answer: \answerYes{}
    \item[] Justification: The original creators are cited, and Appendix~\ref{app:licenses} identifies the upstream code, model, and dataset terms.

\Needspace{11\baselineskip}
\item {\bf New assets}
    \item[] Question: Are new assets introduced in the paper well documented and is the documentation provided alongside the assets?
    \item[] Answer: \answerNA{}
    \item[] Justification: This paper analyzes existing models and datasets and introduces no new datasets or pretrained models.

\Needspace{11\baselineskip}
\item {\bf Crowdsourcing and research with human subjects}
    \item[] Question: For crowdsourcing experiments and research with human subjects, does the paper include the full text of instructions given to participants and screenshots, if applicable, as well as details about compensation (if any)?
    \item[] Answer: \answerNA{}
    \item[] Justification: The reported work uses existing benchmark evaluations and does not involve recruited participants or crowdsourcing.

\Needspace{11\baselineskip}
\item {\bf Institutional review board (IRB) approvals or equivalent for research with human subjects}
    \item[] Question: Does the paper describe potential risks incurred by study participants, whether such risks were disclosed to the subjects, and whether Institutional Review Board (IRB) approvals (or an equivalent approval/review based on the requirements of your country or institution) were obtained?
    \item[] Answer: \answerNA{}
    \item[] Justification: The reported work does not involve human-participant research.

\Needspace{11\baselineskip}
\item {\bf Declaration of LLM usage}
    \item[] Question: Does the paper describe the usage of LLMs if it is an important, original, or non-standard component of the core methods in this research? Note that if the LLM is used only for writing, editing, or formatting purposes and does \emph{not} impact the core methodology, scientific rigor, or originality of the research, declaration is not required.
    \item[] Answer: \answerNA{}
    \item[] Justification: Core method development does not involve LLMs as algorithmic components; the evaluated diffusion MLLMs are the subjects of empirical analysis rather than tools in the proposed methodology. LLM assistance was limited to routine manuscript editing and formatting.

\end{enumerate}



\begin{thebibliography}{99}
\small
\bibitem[Wei et~al.(2022)]{wei2022}
Jason Wei, Xuezhi Wang, Dale Schuurmans, Maarten Bosma, Brian Ichter, Fei Xia, Ed Chi, Quoc V.~Le, and Denny Zhou.
\newblock Chain-of-thought prompting elicits reasoning in large language models.
\newblock In \emph{Advances in Neural Information Processing Systems}, 2022.
\newblock \url{https://arxiv.org/abs/2201.11903}.

\bibitem[Turpin et~al.(2023)]{turpin2023}
Miles Turpin, Julian Michael, Ethan Perez, and Samuel R.~Bowman.
\newblock Language models don't always say what they think: Unfaithful explanations in chain-of-thought prompting.
\newblock In \emph{Advances in Neural Information Processing Systems}, 2023.
\newblock \url{https://arxiv.org/abs/2305.04388}.

\bibitem[Lanham et~al.(2023)]{lanham2023}
Tamera Lanham, Anna Chen, Ansh Radhakrishnan, Benoit Steiner, Carson Denison, Danny Hernandez, et~al.
\newblock Measuring faithfulness in chain-of-thought reasoning.
\newblock arXiv:2307.13702, 2023.

\bibitem[Sahoo et~al.(2024)]{sahoo2024}
Subham Sekhar Sahoo, Marianne Arriola, Yair Schiff, Aaron Gokaslan, Edgar Marroquin, Justin T.~Chiu, Alexander Rush, and Volodymyr Kuleshov.
\newblock Simple and effective masked diffusion language models.
\newblock In \emph{Advances in Neural Information Processing Systems}, 2024.
\newblock \url{https://arxiv.org/abs/2406.07524}.

\bibitem[Li et~al.(2025a)]{li2025prophet}
Pengxiang Li, Yefan Zhou, Dilxat Muhtar, Lu Yin, Shilin Yan, Li Shen, Soroush Vosoughi, and Shiwei Liu.
\newblock Diffusion language models know the answer before decoding.
\newblock arXiv:2508.19982, 2025. Version~5, April 2026.

\bibitem[Yeom et~al.(2026)]{yeom2026}
Jewon Yeom, Jaewon Sok, Seonghyeon Park, Jeongjae Park, Hwiyeong Lee, and Taesup Kim.
\newblock Answer first, reason later: When commitment order costs accuracy in diffusion language models.
\newblock arXiv:2608.05687, 2026. Version~2.

\bibitem[Favero et~al.(2024)]{favero2024}
Alessandro Favero, Luca Zancato, Matthew Trager, Siddharth Choudhary, Pramuditha Perera, Alessandro Achille, Ashwin Swaminathan, and Stefano Soatto.
\newblock Multi-modal hallucination control by visual information grounding.
\newblock In \emph{Proceedings of the IEEE/CVF Conference on Computer Vision and Pattern Recognition}, pages 14303--14312, 2024.
\newblock \url{https://arxiv.org/abs/2403.14003}.

\bibitem[Kim et~al.(2026)]{kim2026thinking}
Keuntae Kim, Mingyu Kang, and Yong Suk Choi.
\newblock Thinking Diffusion: Penalize and guide visual-grounded reasoning in diffusion multimodal language models.
\newblock arXiv:2604.05497, 2026.

\bibitem[Kim et~al.(2026)]{kim2026st}
Keuntae Kim, Beomseok Lee, Hyunwoo Kim, and Yong Suk Choi.
\newblock ST-Veto: Spatio-temporal token veto for diffusion MLLMs via Taylor prediction and visual grounding.
\newblock arXiv:2607.17884, 2026.
\newblock \url{https://arxiv.org/abs/2607.17884}.

\bibitem[Li et~al.(2025b)]{li2025lavida}
Shufan Li, Konstantinos Kallidromitis, Hritik Bansal, Akash Gokul, Yusuke Kato, Kazuki Kozuka, Jason Kuen, Zhe Lin, Kai-Wei Chang, and Aditya Grover.
\newblock LaViDa: A large diffusion language model for multimodal understanding.
\newblock arXiv:2505.16839, 2025.

\bibitem[Fu et~al.(2026)]{fu2026}
Yonggan Fu, Lexington Whalen, Abhinav Garg, Chengyue Wu, Maksim Khadkevich, Nicolai Oswald, Enze Xie, Daniel Egert, et~al.
\newblock Nemotron-Labs-Diffusion: A tri-mode language model unifying autoregressive, diffusion, and self-speculation decoding.
\newblock arXiv:2607.05722, 2026.

\bibitem[DiffusionGemma Team(2026)]{diffusiongemma2026}
DiffusionGemma Team.
\newblock DiffusionGemma technical report.
\newblock arXiv:2608.00146, 2026.

\bibitem[Wu and Xie(2023)]{wu2023}
Penghao Wu and Saining Xie.
\newblock V*: Guided visual search as a core mechanism in multimodal LLMs.
\newblock arXiv:2312.14135, 2023.

\bibitem[Chen et~al.(2024)]{chen2024}
Qiguang Chen, Libo Qin, Jin Zhang, Zhi Chen, Xiao Xu, and Wanxiang Che.
\newblock M$^3$CoT: A novel benchmark for multi-domain multi-step multi-modal chain-of-thought.
\newblock In \emph{Proceedings of ACL}, pages 8199--8221, 2024.
\newblock \url{https://aclanthology.org/2024.acl-long.446/}.

\bibitem[Lu et~al.(2022)]{lu2022}
Pan Lu, Swaroop Mishra, Tony Xia, Liang Qiu, Kai-Wei Chang, Song-Chun Zhu, Oyvind Tafjord, Peter Clark, and Ashwin Kalyan.
\newblock Learn to explain: Multimodal reasoning via thought chains for science question answering.
\newblock In \emph{Advances in Neural Information Processing Systems}, 2022.
\newblock \url{https://arxiv.org/abs/2209.09513}.
\end{thebibliography}
\end{document}